\documentclass[letterpaper, 10 pt, conference]{ieeeconf}  

\IEEEoverridecommandlockouts                              

\usepackage{stdlib}

\title{\LARGE \bf
Iterative Program Synthesis for Adaptable Social Navigation
}

\author{Jarrett Holtz$^{1}$ and Simon Andrews$^{2}$ and Arjun Guha$^{3}$ and Joydeep Biswas$^{1}$
\thanks{$^{1}$University of Texas at Austin.
{\tt\{jaholtz,joydeepb\}@utexas.edu}}%
\thanks{$^{2}$University of Massachusetts, Amherst. {\tt sbandrews@umass.edu}}%
\thanks{$^{3}$Northeastern University. {\tt a.guha@northeastern.edu}}%
}

\begin{document}

\maketitle
\thispagestyle{empty}
\pagestyle{empty}

\begin{abstract}
Robot social navigation is influenced by human preferences and
environment-specific scenarios such as elevators and doors, thus necessitating
end-user adaptability.
State-of-the-art approaches
to social navigation fall into two categories: model-based social
constraints and learning-based approaches.
While effective, these approaches have fundamental limitations --
model-based approaches require constraint and parameter tuning to adapt to
preferences and new scenarios, while learning-based approaches require
 reward functions, significant training data, and are hard to adapt to
new social scenarios or new domains with limited demonstrations.

In this work, we propose Iterative Dimension Informed
Program Synthesis (\technique{}) to address these limitations by learning
and adapting social navigation in the form of human-readable symbolic programs.
\technique{} works by combining program synthesis,
parameter optimization, predicate repair, and iterative human demonstration
to learn and adapt model-free action selection policies from orders of magnitude
less data than learning-based approaches. We introduce a novel predicate repair
technique that can accommodate
previously unseen social scenarios or preferences by growing existing policies.

We present experimental results showing that \technique{}:
1) synthesizes effective policies that model user preference,
2) can adapt existing policies to changing preferences,
3) can extend policies to handle novel social scenarios such as locked doors, and
4) generates policies that can be transferred from simulation to real-world
robots with minimal effort.




\end{abstract}

\section{Introduction}
Social navigation is a fundamental robot behavior that is integrally
tied to human preferences, and that needs to be robust to potentially unknown;
environment-specific decision making.
State-of-the-art approaches to social navigation fall into two major categories,
model-based and model-free approaches.
Model-based approaches include those that employ engineered models of social
constraints, such as the social force model~\cite{socialForce}
or human-robot proxemics~\cite{proxemics}.
However, no single model can capture all possible social constraints, and
adapting models
requires tedious parameter tuning~\cite{costFunctions}.
Model-free approaches include those that leverage Neural Networks (NNs) for
Learning from Demonstration~\cite{burgard2018} or Reinforcement Learning~\cite{how2017}.
However, these approaches suffer from
limitations common to NNs: they are data-intensive~\cite{deepLearningRobots},
difficult to understand~\cite{manuelaExplanation},
and challenging to adapt to new domains without
starting over~\cite{closingSimToReal}.

Recent work on Layered Dimension Informed Program Synthesis (LDIPS)~\cite{holtz2020ldips}
has shown
that program synthesis, when extended with dimensional analysis, addresses many
of these concerns by learning robot behaviors as
human-readable programs.
However, while LDIPS can learn
new behaviors, it does not address policy
adaptation.
In this work, we build upon LDIPS to
present Iterative Dimension Informed Program Synthesis (\technique{})
to synthesize and adapt social navigation behaviors from
human demonstration. A highlight video of our approach can be found at
\url{https://youtu.be/JoT8nZ_Rsto}.

Given a set of demonstrations and an optional
starting behavior, \technique{} produces a new minimally altered behavior
consistent with the demonstrations.
Iterative application of IDIPS allows for continuous refinement
based on further demonstration.
\technique{} employs three
modules to accomplish this:
a synthesis module for learning new behaviors,
a parameter optimization module for adjusting real-numbered parameter values,
and a predicate repair module for adjusting logical components.
For the synthesis module, we extend LDIPS with MaxSMT
constraints to handle potentially conflicting human demonstrations
by allowing for partial satisfaction~\cite{z3}. For the parameter
repair module, we employ SMT-Based Robot Transition Repair
(SRTR)~\cite{holtz2018interactive},
an approach for transition repair based on human corrections.
Finally, for the predicate repair module, we introduce a novel technique
for adapting the conditional logic with the minimal syntactic
changes to accommodate new demonstrations. An open-source implementation
of our approach can be found at \url{https://github.com/ut-amrl/pips}.

We
evaluate \technique{}
in simulation and on real robots
to demonstrate the following:
\begin{inparaenum}
  \item \technique{} can synthesize effective policies
    for social navigation that model the preferences
    of distinct demonstration sets,
  \item synthesized policies can be automatically adapted towards the preferences of
    alternative demonstrations with a small
    number of corrections,
  \item \technique{} can adapt synthesized policies to unseen scenarios, such
    as a locked door, with a small
    number of corrections, and
  \item synthesized programs can be transferred
    to a real-world mobile service robot
    and adapted.
\end{inparaenum}

\section{Background and Related Work}
We frame social navigation as a discounted-reward Markov Decision Process (MDP)
$M = \langle S, A, T, R, \gamma \rangle$ consisting of
the state space $S$ that includes robot, human, and environment states;
actions $A$ represented either as discrete motion primitives~\cite{mpdm2} or
continuous local planning actions~\cite{costFunctions}; the world transition
function
\begin{equation}
  T(s, a, s') = P(s_{t + 1} = s' \given s_t = s, a_t = a)
\end{equation}
for the probabilistic transition to states $s'$ when taking action $a$ at
previous state $s$; the reward function  $R : S \times A \times S \mapsto \real$
; and discount factor $\gamma$. The solution to this MDP is represented as a
policy $\pi : S \times A \mapsto A$ that decides what actions to take
based on the previous state-action pair. The optimal social
navigation policy $\pi^*$ maximizes the expectation over the cumulative discounted rewards
\begin{align}
  \pi^* &= \arg_{\pi}\max J_\pi\nonumber,\\
  J_\pi &= E \left[ \sum_{t=0}^{t=\infty} \gamma^t R(s_t, \pi(s_t, a_t),
  s_{t+1}) \right]
\end{align}
The social navigation problem has been extensively studied, and there are
two broad classes of algorithms -- \emph{model-based} approaches that encode
known models to represent either $R$ or $\pi$ directly, and \emph{model-free}
approaches that do not assume that $R$ is known, and also do not enforce prior
model structure on the learned $\pi$.

\textbf{Model-based approaches} can be further classified based
on which of $\pi$
and $R$ are based on pre-defined models.
Some of the earliest approaches to autonomous navigation in social settings were
with model-based policies, including standing in line~\cite{nakauchi2002social}
and person-following~\cite{gockley2007natural}. Unfortunately such approaches
require explicit enumeration of all possible social scenarios, which is
infeasible in real world settings. To generalize behaviors to novel scenarios,
social factors have been used to model $R$, such as proxemics~\cite{proxemics}
for social navigation~\cite{socialMaps}, or the social
force model~\cite{socialForce,socialForce2}, and learn $\pi$ by optimizing for
the discounted rewards. Such an optimization may be computationally
intractable over continuous state and action spaces -- to overcome this
limitation, multi-policy decision making (MPDM) for social navigation~\cite{mpdm2}
decomposes $M$ into hierarchical policies, where at the highest level, $\pi$
selects lower-level controllers (policies) as actions. MPDM is thus capable of
real-time optimization of the discounted objective function by rollouts of the
hierarchical policy over a receding horizon. However, MPDM still requires
model-based specifications of $R$, which makes it hard to adapt to novel social
circumstances such as taking turns through busy doors, which may not be captured
by the models for $R$.

\textbf{Model-free approaches} rely on function approximators to represent $\pi$
and optionally $R$, most recently using deep neural networks (DNNs).
If $R$ is known, deep reinforcement learning (DRL) can be used to learn a
DNN-based representation of $\pi$~\cite{how2017,hoof2020}.
If $R$ is not known, inverse reinforcement learning (IRL) for social
navigation~\cite{arras2014,arras2016} first infers $R$, and then given access to
$T$ in simulation or via real-world evaluation, learns $\pi$.
If $R$ is neither known, nor easy to infer $\pi$ from, learning for
demonstration (LfD) approaches such as Generative Adversarial Imitation
Learning~\cite{burgard2018} are used to infer $\pi$ directly from state-action
sequences from user demonstrations. Despite their success at inferring $\pi$
DNN-based approaches
suffer from known limitations - they require significant data,
are brittle to domain changes, and once trained are hard to
adapt to new settings.

We propose a novel
solution to social navigation that overcomes limitations of both
model-based, and DNN-based model-free social navigation: synthesis of
$\pi$ as symbolic programs in a model-free LfD setting. IDIPS synthesizes
symbolic policies from orders of magnitude fewer demonstrations than DNN-based
approaches, and since IDIPS does not assume any prior for $R$ or $\pi$, it is
capable of adapting to completely new scenarios.
IDIPS builds on recent work on SMT-based solutions to social
navigation~\cite{campos2019} and robot program repair~\cite{holtz2018interactive},
and extends them using dimension-informed program synthesis~\cite{holtz2020ldips}.

\section{Symbolic Policies for Adaptable \\ Social Navigation}
\seclabel{policy}
We propose learning $\pol$ directly from demonstrations in the form of symbolic
action selection policies (ASPs) written in the language described in
prior work \cite{holtz2020ldips}.
A symbolic ASP represents $\pol$ as a human-readable program where the actions
are discrete subpolicies such as follow, halt, or pass. This formulation
is similar to the MPDM formulation \cite{mpdm2}, where the ASP serves the
role of the reward-based subpolicy selection in MPDM. For the remainder of
this paper, we will refer to these sub-policies as \emph{actions}.

The structure of symbolic ASPs is essential to learning and adapting these
policies.
Consider the structure of the policy shown in \figref{examplePolicy}.
A full \emph{policy} $\pol$ consists of logical branches that each return an
\emph{action} $\action$, each branch consists of a \emph{predicate} $\pred$
that represents decision logic, and is in turn composed of \emph{expressions}
$\expr$
that compute features from elements of $S$ and real-valued \emph{parameters}
$\param$ that determine the decision boundaries.
\begin{figure}[htb!]
  \vspace{-1em}
  \begin{lstlisting}[language=dipsexample]
if (!\tikzmark{a}!$\startAction$==$\ah{\texttt{GoAlone}}$ && $\eh{|p_r-H_p[0]|} > \ph{2.0}$!\tikzmark{b}!): return $\ah{\texttt{GoAlone}}$

elif (!\tikzmark{c}!$\startAction$==$\ah{\texttt{GoAlone}}$ && $\eh{(p_r - H_p[1]).x > \ph{1.0}}$ &&
      $\eh{v_r.x-H_v[0].x} > \ph{0.0}$ $\! \&\&\ \eh{|p_r-H_p[0]|} \leq \ph{2.0}$):!\tikzmark{d}! return $\ah{\texttt{Pass}}$

elif (!\tikzmark{e}!$\startAction$==$\ah{\texttt{GoAlone}}$ && $\eh{(p_r - H_p[1]).x \leq \ph{1.0}}$
      && $\eh{|p_r-H_p[0]| \leq \ph{2.0}}$ ||
      $\eh{v_r.x-H_v[0]} \leq \ph{0.0}$ $\!
      \&\&\ \eh{|p_r-H_p[0]|} \leq \ph{1.0}$):!\tikzmark{f}! return $\ah{\texttt{Follow}}$
\end{lstlisting}
\begin{tikzpicture}[remember picture,overlay]
\draw[orange,rounded corners]
  ([shift={(-3.5pt,1.5ex)}]pic cs:a)
    rectangle
  ([shift={(8.5pt,-0.65ex)}]pic cs:b);
\end{tikzpicture}
\begin{tikzpicture}[remember picture,overlay]
\draw[orange,rounded corners]
  ([shift={(-3.5pt,1.5ex)}]pic cs:c)
    rectangle
  ([shift={(3pt,-0.65ex)}]pic cs:d);
\end{tikzpicture}
\begin{tikzpicture}[remember picture,overlay]
\draw[orange,rounded corners]
  ([shift={(-3.5pt,1.5ex)}]pic cs:e)
    rectangle
  ([shift={(3pt,-0.65ex)}]pic cs:f);
\end{tikzpicture}
  \vspace{-2em}
  \caption{An example symbolic action selection policy (ASP).
  Predicates in each branch are outlined in \bh{orange}. Expressions
  are highlighted in \eh{purple}, parameters in \ph{green}, and actions in \ah{red}.}
\figlabel{examplePolicy}
  \vspace{-1em}
\end{figure}

The structure of the ASPs and the operator rules are described by
the language, but the actions,
elements of $S$, and the library of operators used to calculate $\expr$
are particular to the application domain to which \technique{} is applied.
To learn these policies from user demonstrations we require a way
to demonstrate the action transitions and world states.


\subsection{End-User Policy Demonstrations}
\seclabel{demos}

We propose two methods for leveraging user guidance to
generate demonstrations of the form $\langle a_t, s_t, a_{t+1} \rangle$.
In simulation, we simulate the state transition function $T$ while
executing a policy $\pol$ that continuously executes a fixed
action $a$ to generate a series of demonstrations $\{\langle a, s_t, a\}$.
At any time the user may interrupt the simulation, optionally rewind to a time
$t-n$, and provide a demonstration directing the robot to transition to a
fixed $\pi'$ that executes $a'$.
This process adds a single transition demonstration
$\langle a, s_{t-n}, a'\rangle$,
and continues until the user is satisfied.

In the real world, the robot observes states $s_t$ and continously executes a
policy $\pol$ with the real world providing $T$. At any time, the user may interrupt
the robot and joystick it through a series of states $\{s_{t} \ldots s_{t+n}\}$.
The user then labels this sequence with actions that should have
been executed by identifying the states $s_t$ where the robot
should transition from an action $a_t$ to an action $a_{t+1}$
resulting in a demonstration sequence $\{\langle a_{t-1}, s_t,
a_t \rangle : t\in (t_1, t_{n})]\} $.
\figref{realDemo} shows a visual representation of a demonstration sequence
for a robot moving through a doorway.

\begin{figure}
  \centering
  \includegraphics[width=0.5\columnwidth]{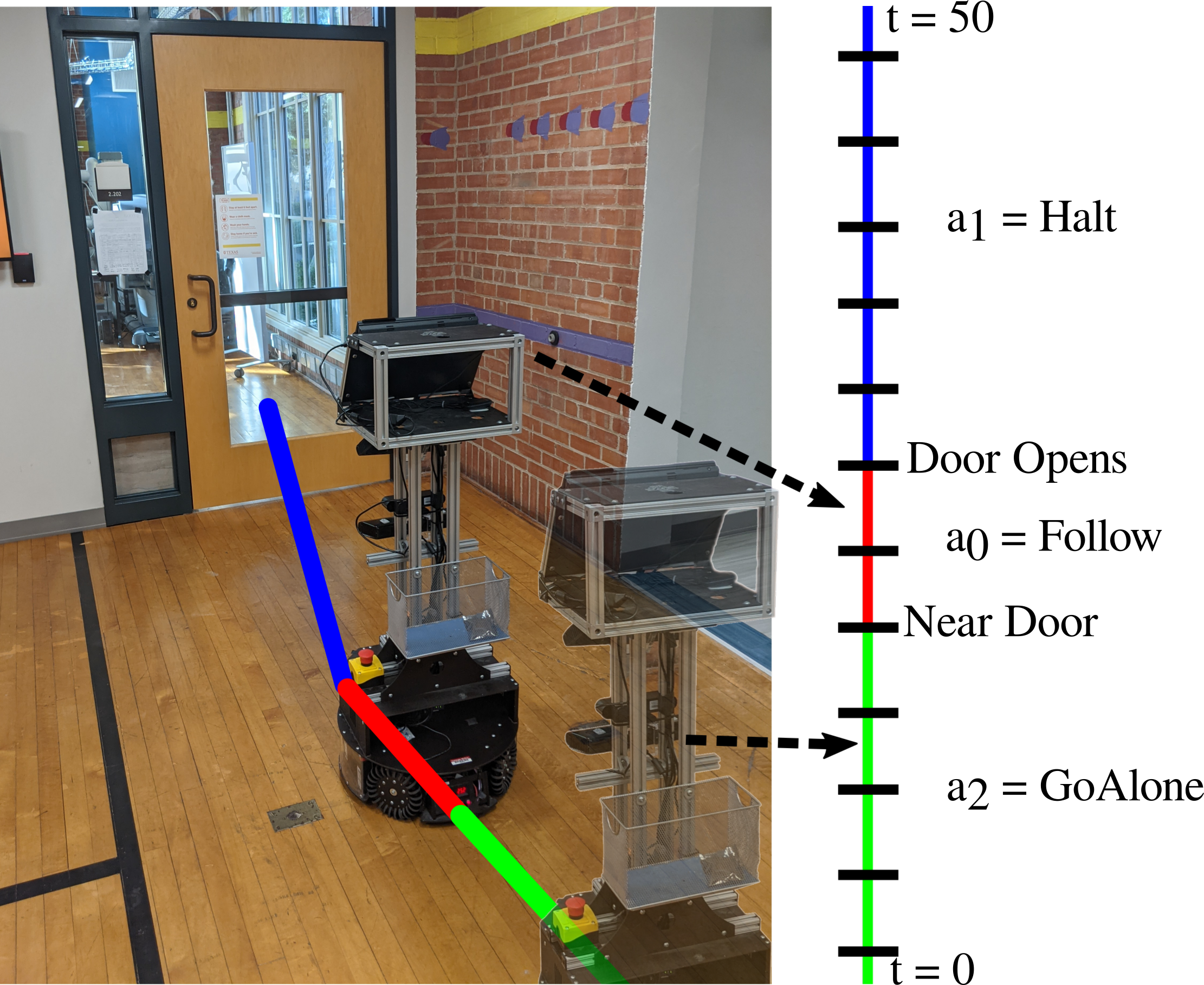}
  \caption{An example labeled demonstration sequence and corresponding timeline
  showing the robot waiting for the door. The overlayed
  timeline is color-coded to show the robot action for that time. }
  \figlabel{realDemo}
  \vspace{-1.5em}
\end{figure}

\section{Iterative Dimension Informed Program Synthesis}

Given a sequence of demonstrations, the goal is to learn a new policy or adapt
an existing one to better match the demonstrations. At a high-level, this
process proceeds as follows:
When there is no initial ASP,
we need to \emph{synthesize} a new symbolic ASP from scratch.
Given an existing policy and a new set of demonstrations,
we identify the components in need of adaptation through \emph{fault localization}.
For each faulty predicate identified by fault localization, we make
minimal modifications to the ASP that improve performance.
Since each predicate is composed of expressions compared to real-valued
parameters, we can optimize performance via \emph{parameter optimization}.
If, after parameter optimization, the policy is still faulty,
then those faults must lie with the predicate itself.
To maximize performance on the new demonstrations while maintaining
performance on prior scenarios, \emph{predicate repair} can be employed
to add new conditional logic.
An end-user can then deploy the ASP, and as adjustments become necessary,
repeat this process to adapt the ASP as shown in \figref{systemDiagram}.

Our approach is Iterative Dimension Informed Program
Synthesis (IDIPS), shown in \figref{metaAlg}.
\technique{} synthesizes policies by combining three modules: a
policy synthesizer for $\pi$ (line~\ref{line:synthesize}), an optimizer
for real-valued parameters (line~\ref{line:optimize}), and a predicate repair
module that extend predicate logic (line~\ref{line:repair}).
\technique{} consumes an optional ASP $\pol$
and a demonstration sequence $\demos$ and produces a new
ASP $\pol'$ that exceeds a minimum success rate $\minScore$ on
$\demos$, if such a $\pol'$ exists.
As our policy synthesis module line~\ref{line:synthesize},
we employ Layered Dimension Informed Program
Synthesis (LDIPS) \cite{holtz2020ldips}.
Given demonstrations, LDIPS produces a symbolic policy matching the structure
described in \secref{policy}
that is constrained by a dimension-informed type system
to prevent synthesis of physically meaningless expressions.
When no initial policy is provided,
then \technique{} is equivalent to running LDIPS alone for the given $\demos$.
When a complete $\pol$ is provided, \technique{} attempts adaptation to satisfy
$\demos$.

\begin{figure}
  \centering
  \includegraphics[width=\columnwidth]{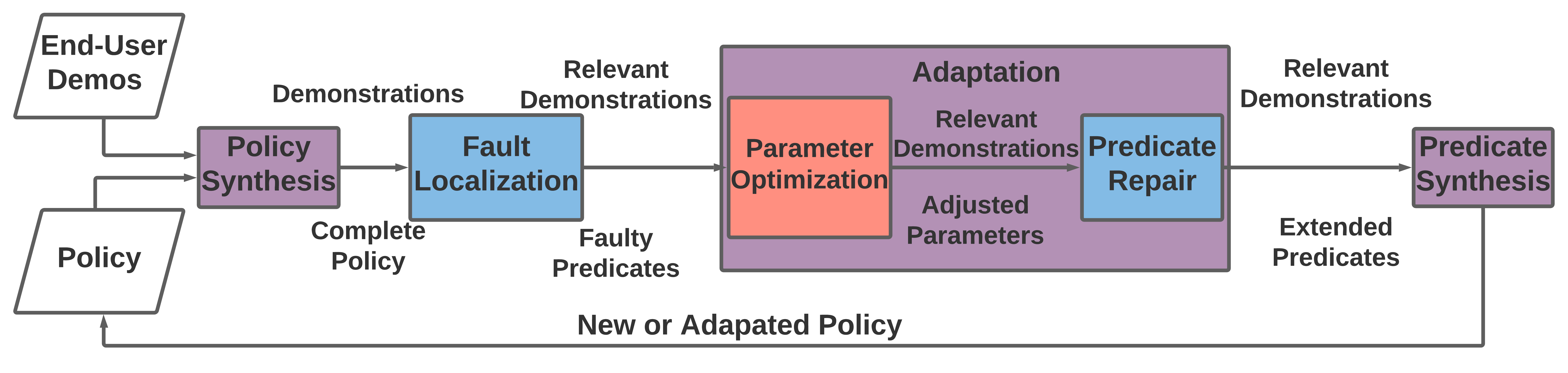}
  \vspace{-2.0em}
  \caption{System diagram of \technique{}.
  Unmodified components are shown in red,
  updated components are shown in purple, and novel components
  are shown in blue.}
  \figlabel{systemDiagram}
  \vspace{-2.0em}
\end{figure}

\begin{figure*}[!tbh]
  \begin{subfigure}[t]{0.3\textwidth}
    \input{idips_fig.tex}
  \end{subfigure}
  \begin{subfigure}[t]{0.4\textwidth}
    \input{paramOptimization.tex}
  \end{subfigure}
  \begin{subfigure}[t]{0.3\textwidth}
    \input{predicateRepair.tex}
  \end{subfigure}
  \vspace{-0.5em}
  \caption{Algorithms for Iterative Dimension Informed Program Synthesis (IDIPS)
  and accompanying submodules.}
  \vspace{-2em}
\end{figure*}

\subsection{Fault Localization}
\technique{} needs only
adapt predicates with suboptimal performance with respect to $\demos$.
We identify these predicates using a \emph{fault localization} procedure
made possible by the structure of our symbolic ASPs.
As shown in \figref{metaAlg},
\texttt{FindPredicates} (line \ref{line:findPred})
simulates running $\pol$ on every demonstration and identifies
the predicates $\pred$ in $\pol$ or implied by the demonstrations,
and the
corresponding positive and negative examples ($\posE, \negE$) for which the
$\pred$ should return \texttt{true} and \texttt{false}, respectively.
For a predicate $b$ that determines whether $\pi$ should transition from
executing previous action $a_1$ to a new action $a_2$, positive and negative
examples are drawn from demonstrations $\demos$ such that
\begin{align}
  \posE &= \left\{\langle s_t, a_t, a_{t+1} \rangle \in \demos : a_t = a_1,  a_{t+1} = a_2\right\}, \\
  \negE &= \left\{\langle s_t, a_t, a_{t+1} \rangle \in \demos : a_t = a_1,  a_{t+1} \neq a_2\right\}.
\end{align}

Then, for each $\langle \pred, \posE, \negE \rangle$ the helper function
\texttt{Score} calculates the percentage of examples with which $\pred$
is consistent, and any predicate with a score less than $\minScore$ is
considered faulty (lines \ref{line:optimize} and \ref{line:repair}).
Given a faulty $\pred$, adaptation proceeds with the
optimization module first, and then the repair module if the optimized policy
does not meet $\minScore$.

\subsection{Parameter Optimization}
\seclabel{optimization}
The optimization module adjusts real-valued parameters
to match the demonstrated behavior.
We employ SMT-based Robot Transition Repair (SRTR) as our optimization module
\cite{holtz2018interactive}. SRTR is a white-box approach
that leverages MaxSMT to correct parameters based on user-supplied
corrections.
A high-level version of the SRTR algorithm is shown in \figref{paramSynth}.
In this algorithm, the helper function $\texttt{ExtractParams}$
(line~\ref{line:ExtractParams}) identifies all parameters in a predicate via
program analysis.
Given $\vec{\param}$
(line \ref{line:startSMT}) employs a canonical partial
evaluator~\cite{partialEval} to form
a simplified representation of $\posE$, $\negE$, $\pred$, and $\vec{\param}$
as a MaxSMT formula, that can be optimized with an off the
shelf MaxSMT solver~\cite{z3}.
By leveraging SRTR's MaxSMT approach, we can satisfy the
maximum number of new demonstrations while minimizing changes to the parameters,
yielding updated predicates $\pred'$ that can be incorporated into
$\pol'$ and are amenable to further repair.

\subsection{Predicate Repair}
\seclabel{dipr}
Parameter optimization and synthesis
are not always sufficient
for adapting policies -- parameter optimization cannot add new
conditional branches in an ASP, and from-scratch synthesis risks degrading
performance on \emph{previously working}
demonstrations. To overcome these challenges, IDIPS uses
predicate repair to extend conditionals while minimizing syntactic
change.
Predicate repair consumes a
predicate $\pred$ and sets of positive and negative examples
$\posE$ and $\negE$ for which $\pred$ has been determined to be faulty, and
outputs an extended predicate $\pred'$ with
minimal syntactic extensions and improved performance on
$\posE, \negE$.

The predicate repair algorithm (\figref{predRepair}) proceeds in three steps:
fault classification, predicate extension, and predicate synthesis.
\begin{inparaenum}
  \item Fault classification identifies whether $\pred$ returns
    false positives, false
    negatives, or both, given the example sets $\posE$ and
    $\negE$ (lines~\ref{line:falseNeg}--\ref{line:falsePos}).
  \item Predicate extension (lines~\ref{line:fullExt}--\ref{line:andExt})
    builds on $\pred$ with three kinds of predicate placeholders $\pred'$,
    that either weaken, strengthen,
    or both.
    If $\pred$ contains both false positives and false negatives, $\pred$ is
    extended using both a conjunction and a disjunction to produce
    $\pred'$ (line~\ref{line:fullExt}). If the only failures
    are false negatives, $\pred$ is made less strict using a disjunction to yield
    $\pred'$ (line~\ref{line:orExt}). If the only failures are false positives,
    $\pred$ is made stricter using a conjunction, to produce
    $\pred'$ (line~\ref{line:andExt}).
  \item Predicate synthesis completes the parts of $b'$ that were
    extended -- $\ppred$ and $\ppred'$ (line~\ref{line:predSynth}) by
    synthesizing new structure, expressions, and parameters to
    replace them.
\end{inparaenum}

\subsection{Predicate Synthesis}
\seclabel{predSynthesis}
Given an extended predicate $\pred'$ predicate synthesis completes placeholder
$\ppred$ such that $\pred'$ has maximal
performance on $\posE$ and $\negE$. Predicate synthesis employs a
fragment of LDIPs~\cite{holtz2020ldips} extended to handle
conflicting demonstrations using MaxSMT. This fragment of LDIPS is
responsible for enumerating and completing all candidate predicates.
Internally predicate
synthesis requires expression synthesis to complete candidate predicates
with physically meaningful expressions.
Expression synthesis employs feature enumeration
to iteratively synthesize all possible expressions.
In turn, expression synthesis employs parameter synthesis
to find real-valued threshold parameters that satisfy all demonstrations, if
any such parameters exist for a candidate program.

We extend this fragment of LDIPS with MaxSMT constraints by
using parameter optimization as in ~\secref{optimization}. Instead of searching
for parameter values that satisfy all demonstrations
we solve for values that satisfy the maximum subset
of the demonstrations. Then, for each of expression synthesis and predicate
synthesis, we consider all possible candidates and return the best
performing solution.
When predicate synthesis is complete, the best performing adapted policy $\pol'$
consisting of the best performing $\pred'$
is returned. This process can continue iteratively, with the user deploying
$\pol'$ and providing new demonstrations to $\technique{}$ as necessary.


\section{Evaluation}
We present results from several experiments in the social navigation domain
that evaluate \technique{'s} ability to
\begin{inparaenum}
\item effectively synthesize policies that reflect user preferences,
\item adapt existing policies to different preferences,
\item repair existing policies for novel scenarios, and
\item transfer programs to real-world robots and repair for performance.
\end{inparaenum}

\subsection{Implementation}
We consider a mobile service robot deployed in the hallways of a
busy office building
moving alongside humans.
For simulating humans, we
employ pedsim\_ros~\footnote{\href{https://github.com/srl-freiburg/pedsim_ros}{https://github.com/srl-freiburg/pedsim\_ros}}~, a crowd simulator built on libpedsim.
We use a virtual hallway with three
possible starting and ending robot locations for our simulated
experiments, as shown in~\figref{simEnviron}.

We define the world state $w = \langle S_r, S_h, S_e \rangle$ to include the
robot state $S_r$ consisting of global and local goal locations,
human states $(S_h)$ consisting of the poses and velocities of the three closest
humans, and the state of navigation relevant
elements of the environment ($S_e$), such as the map and door state.
Our ASPs consist of four actions, stopping in place
(Halt), navigating to the goal (GoAlone), following a human (Follow),
and passing a human (Pass).

\begin{figure}[htb!]
  \vspace{-1em}
  \centering
  \includegraphics[width=0.9\columnwidth]{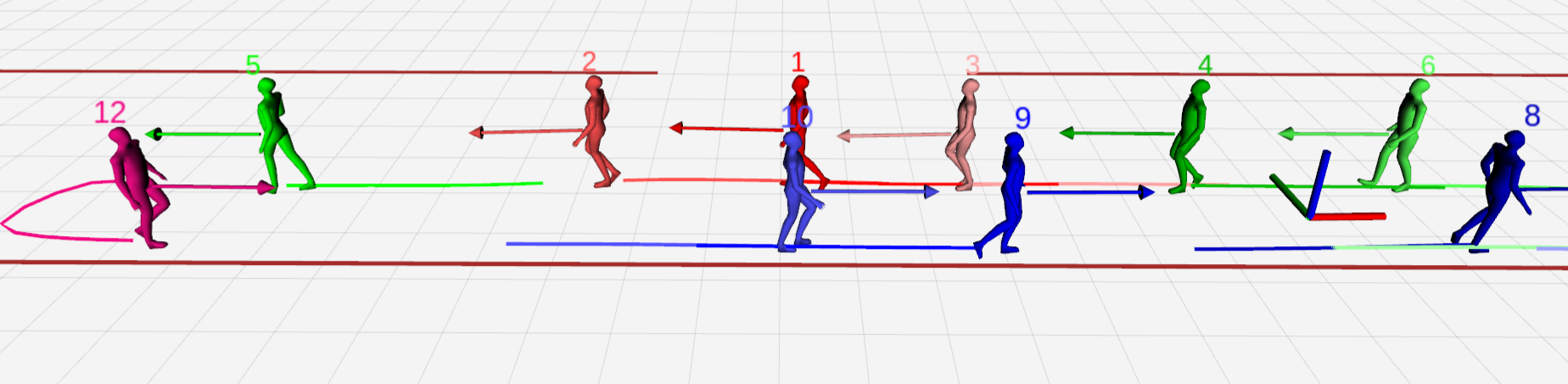}
  \vspace{-0.5em}
  \caption{Example Simulation Environment.}
  \figlabel{simEnviron}
  \vspace{-2em}
\end{figure}

\begin{figure*}[htb!]
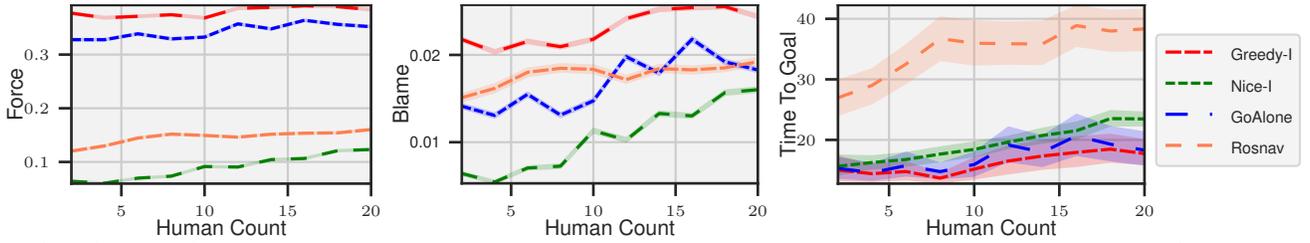

  \centering
  \hspace{-0.1\textwidth}
  \input{images/simForce.pgf}
  \input{images/simBlame.pgf}
  \begin{subfigure}[t]{0.3\textwidth}
  \input{images/simTime.pgf}
  \end{subfigure}
  \vspace{-0.5em}
  \caption{Performance for synthesized policies and baselines.
  The shaded regions regions around lines represent the 90\% confidence interval.}
  \figlabel{metrics1}
  \vspace{-0.5em}
\end{figure*}
\begin{figure*}[hbt!]
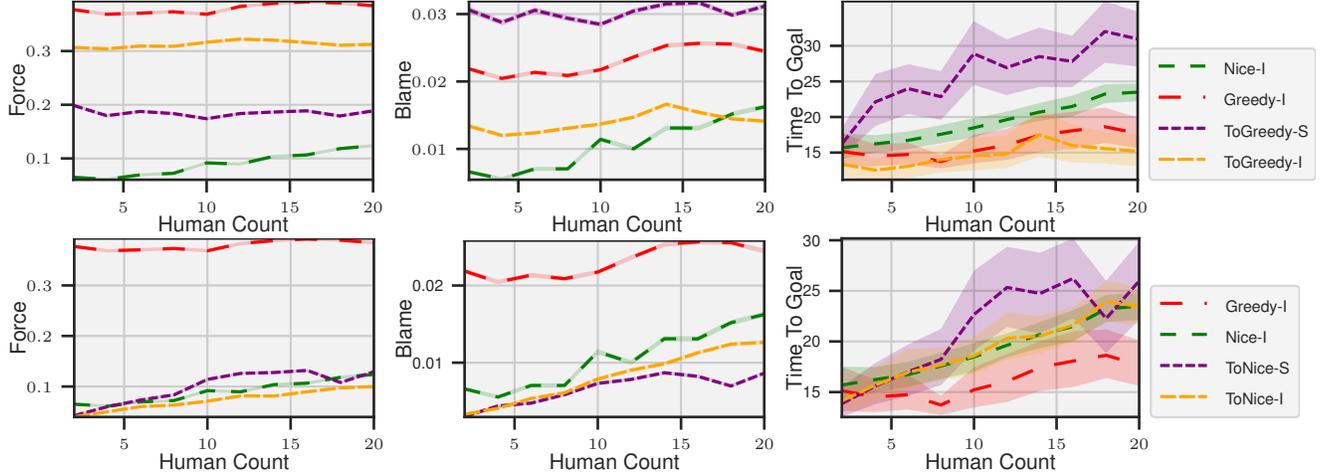

  \centering
  \hspace{-0.1\textwidth}
  \input{images/adaptNF.pgf}
  \input{images/adaptNB.pgf}
  \begin{subfigure}[t]{0.3\textwidth}
  \input{images/adaptedNiceTime.pgf}
  \end{subfigure}\\
  \hspace{-0.1\textwidth}
  \input{images/adaptGF.pgf}
  \input{images/adaptGB.pgf}
  \begin{subfigure}[t]{0.3\textwidth}
  \input{images/adaptedGreedyTime.pgf}
  \end{subfigure}
  \caption{Performance for adapted policies.
  The shaded regions regions around lines represent the 90\% confidence interval.}
  \figlabel{metrics2}
  \vspace{-0.5em}
\end{figure*}
\begin{figure*}[htb!]
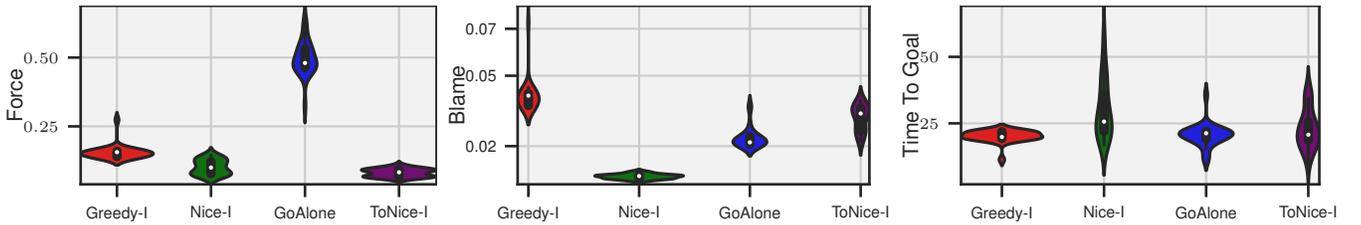

  \centering
  \input{images/realForce.pgf}
  \input{images/realBlame.pgf}
  \input{images/realTime.pgf}
  \caption{Performance from real robot experiments.}
  \figlabel{metrics4}
  \vspace{-1.5em}
\end{figure*}

\subsection{Metrics}
As a model-free approach to social navigation, LDIPS does not need any reward
function to be specified -- it directly synthesizes ASPs from user
demonstrations. However, we use three quantitative metrics to measure the
difference in performance between different ASPs. To evaluate the impact on
humans during navigation, we evaluate two social metrics,
force and blame, as described in work on MPDM~\cite{mpdm2}.
Force is a metric of distance between the robot and the closest human, and
represents the influence humans and the robot exert on each
other as they travel, while blame is a metric of position and velocity that
approximates the robot's level of responsibility
for this influence. To evaluate efficiency,
we report the time in seconds to reach the goal.


\subsection{Performance of Synthesized Social Navigation}
To evaluate the performance of \technique{} synthesized ASPs, we synthesize
policies from two demonstration sets (Nice, Greedy) with opposite preferences, a
\textit{Nice-I} policy that prioritizes socially
passive behavior, and a \textit{Greedy-I} behavior
that prioritizes time to goal. The demonstration sets were human-generated
in simulation as described in \secref{demos}, based on user interpretations
of "Nice" and "Greedy".
We compare synthesized policies to
a GoAlone policy representing
the default behavior of our robot using only the GoAlone action
and a solution leveraging the ROS navigation stack extended with
social
awareness that we refer to as Rosnav~\footnote{\href{http://wiki.ros.org/social\_navigation\_layers}{http://wiki.ros.org/social\_navigation\_layers}}.

\begin{figure}[htb]
  \begin{center}
  \footnotesize
\renewcommand{\arraystretch}{0.8}
    \begin{tabular}{lccc}\toprule
      \multirow{2}{*}{\textbf{Demo Set}} &
      \multicolumn{2}{c}{\textbf{Satisfied ($\%$)}} &
      \multirow{2}{*}{\# of Demo Timesteps}  \\
      \cmidrule(lr){2-3}
                 & \textbf{Nice-I} & \textbf{Greedy-I} & \\
      \midrule
      Nice       & 92 & 76 & 5642 \\
      \midrule
      Greedy     & 81 & 94 & 2618  \\ \bottomrule
    \end{tabular}
  \vspace{-.5em}
  \caption{Percentage of demonstration sets satisfied per policy.}
  \figlabel{demoTable}
  \end{center}
  \vspace{-1.5em}
\end{figure}

We vary the number of humans and robot goal
during $25$ trials each
for a total of $1000$ trials for each policy, and
present the results in \figref{metrics1}.
Greedy-I consistently reaches
the target in the shortest time but also exhibits the highest force and blame.
In comparison, Nice-I is slower but exhibits the lowest force and
blame. While Rosnav achieves comparable performance to
Nice-I in terms of Force and Blame, it comes with a significant increase in
travel time.

To evaluate how closely the synthesized behaviors model their respective
demonstration sets, we show the number of demonstrations from each demo
set satisfied by each synthesized policy in \figref{demoTable}. As expected,
this table shows that synthesized behaviors most closely model the demonstrations
from which they were synthesized.

\subsection{Adapting Social Navigation Preferences}
To simulate a scenario where an end-user wants to adapt an existing policy
for their preferences or domain, we adapt Nice-I with
demonstrations from Greedy-I, and vice versa using two methods:
parameter optimization (SRTR)~\cite{holtz2018interactive} (\textit{ToGreedy-S, ToNice-S}), and \technique{}
(\textit{ToGreedy-I, ToNice-I}). We repeat our experimental procedure
for these policies and report the results in \figref{metrics2}.
For blame and force, the performance shifts as expected for both SRTR and
\technique{} -- the ToNice-* ASPs sacrifice time efficiency for social niceness,
while the ToGreedy-* ASPs prioritize time efficiency over social niceness.
However, the adaptation is significantly more effective using \technique{}
than using only SRTR, validating our hypothesis that program repair is key to
improving adaptation beyond parameter optimization -- ToGreedy-S is
unable to improve
navigation time as significantly as ToGreedy-I.
Similarly, ToNice-S is unable to navigate as efficiently as ToNice-I, and
thus has higher time to goal.

\subsection{Real World Evaluation}
To evaluate transferring \technique{} policies from simulation to the real world,
we employ a Cobot~\cite{cobot} mobile service robot.
Per COVID-19 safety
guidelines, only a single human participant was used for this case study.
We consider two scenarios in this environment and
four policies transferred from simulation
(GoAlone, Nice-I, Greedy-I, ToNice-I).
For each policy, we vary the travel direction and speed of the human and the
starting location of the robot for a total of $25$ trials per policy.
For the results shown in \figref{metrics4},
all of the policies are able to accomplish the
goal while roughly maintaining their relationship in terms of our metrics. The
Nice-I behavior is still the most passive policy, while the Greedy-I
behavior is more aggressive, showing that repair for domain transfer with \technique{}
is effective.



\subsection{Adapting To Unseen Scenarios}
To evaluate \technique{'s} adaption to novel scenarios, we
performed two sets of experiments by adding a closed
door in both simulation and a real hallway~\figref{realDemo}.
%
We evaluate the Nice-I and Greedy-I policies, and those same
policies
adapted using demonstrations of how to wait for the door
before proceeding (\textit{NiceDoor-I, GreedyDoor-I}). We adapt the policies
separately for simulated and real-world examples and record
$200$ trials with varying human counts in simulation
and ten trials with a single human
in the real-world. \figref{doorTable} shows the percentage of successful
runs for each policy.
Neither original policy waits for the door, leading to almost complete failure. In contrast, the repaired policies are able to handle the
new scenarios with minimal failures in simulation and complete success
in the real-world.
This adaptation would require
extensive modification for Rosnav and is only
possible when relearning from scratch in DNN-based approaches.

\begin{figure}[htb]
  \vspace{-1.0em}
  \begin{center}
  \footnotesize
\renewcommand{\arraystretch}{0.5}
    \begin{tabular}{lcc }
      \toprule
      \multirow{2}{*}{\textbf{Policy}} &
      \multicolumn{2}{c}{\textbf{Success Rates ($\%$)}}  \\
      \cmidrule(lr){2-3}
                 & \textbf{Simulation} & \textbf{Real World}\\
      \midrule
      Nice-I       & 2 & 0 \\
      \midrule
      Greedy-I     & 7 & 0  \\
      \midrule
      NiceDoor-I   & 100 & 100  \\
      \midrule
      GreedyDoor-I & 84 & 100 \\
      \bottomrule
    \end{tabular}
  \vspace{-.5em}
  \caption{Success rates for different ASPs on closed door scenarios.}
  \figlabel{doorTable}
  \end{center}
  \vspace{-2.5em}
\end{figure}

%

\section{Conclusion}
In this work, we presented an approach for learning and adapting
symbolic social navigation policies that builds on
dimension informed program synthesis.
We model social navigation as an action selection problem and
learn and adapt behaviors with small numbers of human-generated demonstrations.
Our experimental evaluation demonstrated that this technique can learn effective
social navigation policies that model the preferences of the demonstrations
and that these symbolic policies can be efficiently adapted for
changing user preference and novel scenarios. Further, we demonstrated
in a case study that our technique can adapt policies
on real-world mobile service robots.

\section{Acknowledgments}
This work is conducted in collaboration between the AMRL at
UT Austin and Professor Arjun Guha at Northeastern University, and is supported in part by NSF (CAREER-2046955,
IIS-1954778, SHF-2006404, CCF-2102291 and CCF-2006404), ARO (W911NF-19-2-0333), DARPA (HR001120C0031),
Amazon, JP Morgan, and Northrop Grumman Mission Systems.

\bibliographystyle{IEEEtran}
\bibliography{IEEEabrv,references}

\end{document}